\documentclass[sigconf]{aamas}

\usepackage{balance} % for balancing columns on the final page
\usepackage{graphicx}
\usepackage{textcomp}
\usepackage{xcolor}
\usepackage{graphicx}
\usepackage{algpseudocode}
\usepackage{multirow} 
\usepackage{amsmath}
\usepackage{enumitem}
\setlist{nosep}
\usepackage{caption}
\usepackage{subcaption}
\usepackage{multicol}
\usepackage{adjustbox}
\usepackage{algorithm}
\usepackage{algpseudocode}
\makeatletter
\gdef\@copyrightpermission{
  \begin{minipage}{0.2\columnwidth}
   \href{https://creativecommons.org/licenses/by/4.0/}{\includegraphics[width=0.90\textwidth]{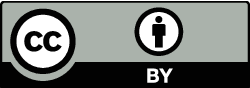}}
  \end{minipage}\hfill
  \begin{minipage}{0.8\columnwidth}
   \href{https://creativecommons.org/licenses/by/4.0/}{This work is licensed under a Creative Commons Attribution International 4.0 License.}
  \end{minipage}
  \vspace{5pt}
}
\makeatother

\setcopyright{ifaamas}
\acmConference[AAMAS '25]{Proc.\@ of the 24th International Conference
on Autonomous Agents and Multiagent Systems (AAMAS 2025)}{May 19 -- 23, 2025}
{Detroit, Michigan, USA}{Y.~Vorobeychik, S.~Das, A.~Nowé  (eds.)}
\copyrightyear{2025}
\acmYear{2025}
\acmDOI{}
\acmPrice{}
\acmISBN{}

\acmSubmissionID{<<932>>}

\title[AAMAS-2025 Formatting Instructions]{Coherence-Driven Multimodal Safety Dialogue with Active Learning for Embodied Agents}

\author{Sabit Hassan}
\affiliation{
  \institution{University of Pittsburgh}
  \city{Pittsburgh, PA}
  \country{USA}}
\email{sabit.hassan@pitt.edu}

\author{Hye-Young Chung}
\affiliation{
  \institution{Northeastern University}
  \city{Boston, MA}
  \country{USA}}
\email{chung.hyey@northeastern.edu}

\author{Xiang Zhi Tan}
\affiliation{
  \institution{Northeastern University}
  \city{Boston, MA}
  \country{USA}}
\email{zhi.tan@northeastern.edu}

\author{Malihe Alikhani}
\affiliation{
  \institution{Northeastern University}
  \city{Boston, MA}
  \country{USA}}
\email{m.alikhani@northeastern.edu}

\begin{abstract}
When assisting people in daily tasks, robots need to accurately interpret visual cues and respond effectively in diverse safety-critical situations, such as sharp objects on the floor. In this context, we present \textbf{M-CoDAL}, a multimodal-dialogue system specifically designed for embodied agents to better understand and communicate in safety-critical situations. The system leverages discourse coherence relations to enhance its contextual understanding and communication abilities. To train this system, we introduce a novel clustering-based active learning mechanism that utilizes an external Large Language Model (LLM) to identify informative instances. Our approach is evaluated using a newly created multimodal dataset comprising \textbf{1K} safety violations extracted from \textbf{2K} Reddit images. These violations are annotated using a Large Multimodal Model (LMM) and verified by human annotators. Results with this dataset demonstrate that our approach improves resolution of safety situations, user sentiment, as well as safety of the conversation. Next, we deploy our dialogue system on a Hello Robot Stretch robot and conduct a within-subject user study with real-world participants. In the study, participants role-play two safety scenarios with different levels of severity with the robot and receive interventions from our model and a baseline system powered by OpenAI's ChatGPT. The study results corroborate and extend the findings from the automated evaluation, showing that our proposed system is more persuasive in a real-world embodied agent setting.
\end{abstract}


\newcommand{\BibTeX}{\rm B\kern-.05em{\sc i\kern-.025em b}\kern-.08em\TeX}

\begin{document}

%%% The following commands remove the headers in your paper. For final 
%%% papers, these will be inserted during the pagination process.

\pagestyle{fancy}
\fancyhead{}

%%% The next command prints the information defined in the preamble.

\maketitle 

%%%%%%%%%%%%%%%%%%%%%%%%%%%%%%%%%%%%%%%%%%%%%%%%%%%%%%%%%%%%%%%%%%%%%%%%

\section{Introduction}

Embodied agents, such as assistive robots with multimodal capabilities, are poised to become an integral part of our daily lives by helping with household tasks and providing assistance upon request.
However, the agent should also play a proactive role and ensure both its and the user's actions are safe. 
It needs to not only detect unsafe scenarios but also communicate and persuade the user of the danger.
To achieve this, the system needs to understand and adapt to conversational contexts while remaining robust across diverse safety-related scenarios.

To meet this challenge, we integrate theories of coherence relations to enhance contextual understanding and use clustering-based active learning to train a multimodal dialogue system, \textbf{M-CoDAL}, specifically designed for deploying embodied agents for real-world users to provide effective safety advice.
\begin{figure*}
    \begin{centering}
  \includegraphics[width=0.7\textwidth]{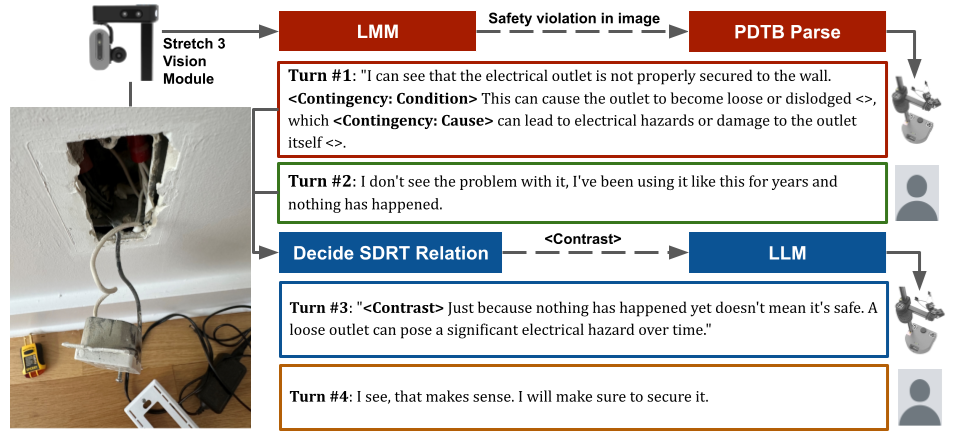}
  \caption{Our proposed dialogue system first parses safety violation in an image with Penn Discourse Treebank \citep{Prasad2008ThePD} relations, and then generate an appropriate response by choosing an Segmented Discourse Representation Theory relation \citep{Asher2005LogicsOC}.}
  \label{fig:discourse-dialogue}
  \end{centering}
\end{figure*}

\textbf{M-CoDAL} uses discourse coherence relations to enrich its contextual understanding. Theories of coherence relations originate in understanding and analyzing inferential links to support text interpretation and has subsequently been extended to cross-modal settings \citep{Alikhani2023ImagetextCA}. Coherence relations can situate an ongoing scene in the arc of a narrative \citep{Cohn2013VisualNS} or enrich interpretation of communicative actions across modalities \citep{Lascarides2009DiscourseCA}. For instance, the coherence relation \textit{Cause} can help interpret why paying attention to a safety violation is important in an image captured by a robot given a certain \textit{Condition} relation. These relations facilitate dialogue between the human user and the robot. By default, a robot equipped with pretrained models would not explicitly model such inferential links \citep{Alikhani2023ImagetextCA}. 
We hypothesize that embodied agents that parse coherence relations of safety violation in captured images and use them to guide their responses will better understand contexts of safety and engage in safer conversations with humans.
%We hypothesize that enabling robots to parse coherence relations of safety violation in captured images, and to use these relations to guide their responses, will result in embodied agents that better understand contexts of safety and engage in safer conversations with humans.

% We hypothesize that parsing coherence relations of safety violations in images that a robot may capture, along with coherence relations to guide the robot's response will result in embodied agents that are more robust in understanding context of safety situations and hold safer conversations with the human user. 
To train\textbf{ M-CoDAL} for embodied agents with coherence relation integration, we adopt clustering-based active learning \citep{Settles2009ActiveLL}. Active learning offers advantage over standard fine-tuning methods by focusing resources toward informative instances. When combined with clustering, active learning has been shown to yield more representative models \citep{hassan-alikhani-2023-calm}. Thus, we hypothesize that applying clustering-based active learning would lead to better coverage of safety scenarios and a safer multimodal dialogue system. Active learning, however, has mostly been addressed in the context of classification tasks and has remained challenging for generative tasks \citep{perlitz2023active} such as conversational response generation due to difficulty in estimating model uncertainty in a large output space. We make the first inroads in active learning for autonomous agents with conversational capacity by integrating an external LLM that quantifies informativeness of an instance based on a composite score of safety of conversation, resolution, and user sentiment. Further, we distill knowledge from another external LLM to reduce necessary human efforts in active learning. Our work is also one of the first to deploy and evaluate active learning in practice for human-robot interactions.

% Recently, \citep{} proposed a mechanism for effective active learning utilizing an auxiliary model that transforms interim output to one dimensional space. In this work, we expand on the work of \citep{} to integrate a more generalizable auxiliary model that maps the interim output conditioned on multiple attributes of the dialogue.

To train our system and evaluate our approach, we first construct a novel dataset consisting of \textbf{1K} safety violations obtained from \textbf{2K} Reddit images. The safety violations are obtained using a Large Multimodal Model (LMM) and then verified, or edited if necessary, by human annotators. Coherence relations and appropriate responses based on the coherence relations are distilled from GPT-4 \citep{2023GPT4VisionSC} for these safety violations and used to train \textbf{M-CoDAL}, based on a smaller LLM, Mistral-7B \citep{Jiang2023Mistral7}, in an active learning setting. Our automated evaluation demonstrates that integrating coherence relations leads to higher safety scores, further improved by clustering-based active learning. We also observe that the improvements translate to models such as Llama 3 \citep{dubey2024llama} and Qwen \citep{qwen} that were not part of the original active learning loop.

Finally, we deploy our multimodal dialogue system on a \textbf{Hello Robot Stretch} robot and conduct a within-subject in-person user study where participants interact with our robot in staged unsafety scenarios. 8 participants interacted with the robot operating with GPT-4o or our proposed system, \textbf{M-CoDAL} in low and high severity fake unsafe scenarios. The study results corroborate findings of automated evaluation by showing that participants found our proposed system \textbf{M-CoDAL} to be more persuasive in both conditions. 
Our qualitative analysis of interviews with the participants also reveals that the participants find \textbf{M-CoDAL} to be more attentive to safety situation and can help the user to be more aware of safety situations. Thus, the key contributions of this paper are:

% The study results corroborate findings of the automated metrics show that our proposed model leads to safer responses in adversarial situations over baseline models such as GPT-4o.

\begin{itemize}
    \item A first-of-its-kind publicly available dataset \footnote{\url{https://github.com/sabithsn/multimodal-embodied-safety}} of multimodal dialogues of safety for embodied agents.
    \item A multimodal dialogue system, \textbf{M-CoDAL} that better understands context of safety via coherence relations.
    \item Extension of active learning paradigm for conversational embodied agent with integration of an external LLM.
    \item Deployment of \textbf{M-CoDAL} with a Hello Robot Stretch robot, accompanied with findings of a real-world user study.
\end{itemize}

\section{Related Work}
\paragraph{Safety in Automated Agents} 
There has been a growing interest in development of multimodal dialogue systems in recent years \citep{sun-etal-2022-multimodal,lee-etal-2023-framework,sicilia2023isabel}, especially with the advent of large multimodal models such as GPT-4V \citep{2023GPT4VisionSC} and LLaVA \citep{Liu2023LLaVAPlusLT}. Discussions of safety however, has primarily focused on textual domain \citep{sun-etal-2022-safety,Weidinger2022TaxonomyOR,dinan-etal-2022-safetykit,atwell-etal-2022-appdia} through means of text-classification or integrating guardrails within LLMs \cite{dubey2024llama}. These approaches may not be sufficient to tackle safety situations that may arise in multimodal household scenarios, particularly when the context demands further probing. To our knowledge, our work is the first public work to process visual cues of safety in multimodal dialogues with an embodied agent.

% And very recently, there have also been development of large multimodal models such as GPT-4V \citep{2023GPT4VisionSC} and LLaVA \citep{Liu2023LLaVAPlusLT}. These models are capable of wide-variety of visual-language tasks.

% Our proposed work will also be the first to integrate clustering-based active learning enhanced and knowledge distillation, along with theories of discourse in multimodal conversational AI.

\paragraph{Coherence Relations for Contextual Understanding}
Coherence relations have been proposed as a possible method for controlling generative models and have been shown to aid tasks such as extractive and abstractive summarization \citep{Cohan2018ADA,Xu2020DiscourseAwareNE}. Coherence relation-aware models have also been shown to generate more coherent texts \citep{bosselut-etal-2018-discourse} within a reinforcement learning setting. \citet{alikhani-etal-2020-cross} extend standard text-based discourse relations to cross-modal scenarios. We build on theories of coherence relations to capture the context of safety scenarios and control responses accordingly within a multimodal dialogue system.

\paragraph{Active Learning with Natural Language}
Active learning is a prominent area in machine learning \citep{Settles2009ActiveLL}, and has gained recent attention for tasks involving natural language \citep{zhang-etal-2022-survey} such as intent classification, sentence matching, and named entity recognition \citep{Zhang2019ensemble,bai-etal-2020-pre,Liu2022LTPAN}. 
% Innovations include continued pretraining on unlabeled data and multi-task adaptation \citep{margatina-etal-2022-importance,rotman-reichart-2022-multi}.
However, active learning has predominantly been focused on classification tasks. Recent works targeting LLMs \citep{diao2023active,margatina-etal-2023-active,hassan2024activelearningrobustrepresentative} also focus on tasks with fixed sets of outputs, leaving active learning for generative tasks largely unexplored \citep{perlitz2023active,hassan2024activelearningframeworkinclusive}. Our work is among the first to pioneer active learning for conversational generative tasks and also extend to multimodal scenarios. Further, we deploy and evaluate active learning in practice, whereas prior work have primarily relied on simulations \citep{zhang-etal-2022-survey}.
% Active learning is a prominent area in machine learning \citep{Settles2009ActiveLL}, receiving increased attention in Natural Language Processing \citep{zhang-etal-2022-survey}. Recent applications include active learning with BERT for tasks like intent classification \citep{Zhang2019ensemble}, sentence matching \citep{bai-etal-2020-pre}, and named entity recognition \citep{Liu2022LTPAN}. Innovations include continued pretraining on unlabeled data \citep{margatina-etal-2022-importance} and adaptation to multi-task scenarios \citep{rotman-reichart-2022-multi}. Empirical studies assess active learning strategies on binary classification \citep{ein-dor-etal-2020-active}. Clustering and advanced active learning strategies are also explored \citep{yuan-etal-2020-cold, margatina-etal-2021-active}. Active learning however, has primarily focused on classification tasks. Recent efforts in introducing LLMs with active learning \citep{diao2023active,margatina-etal-2023-active}, assume a fixed set of outputs \textemdash still classification tasks, albeit with LLMs. Active learning for \textit{generative tasks} remains mostly unexplored. \citep{perlitz2023active} show that standard active learning methods for classification do not translate well to generative tasks and call for further investigation. Ours is one of the earliest works to apply active learning for generative tasks, as well as a multimodal scenario. Our work is also one of the first to deploy and evaluate active learning in practice \citep{zhang-etal-2022-survey}.

\paragraph{Embodied agents with AI models} With advancement in language models and multimodal models, there has been a surge in research to integrate these models with embodied agents \citep{ahn2022icanisay,driess2023palmeembodiedmultimodallanguage,huang2022innermonologueembodiedreasoning,huang2022languagemodelszeroshotplanners}. Most of these recent works focus on extracting plans or executable actions by the robot from pre-trained language models. In contrast, our work enables the robot to observe safety scenarios in its surroundings and communicate with the user. Our real-world user study, conducted by enacting scenarios letting the users interact with the robot, also reveals insights on how human users may perceive such an agent.

% Different from the aforementioned works, ours is the first to develop a multimodal dialogue system for safety in embodied agents and conduct a real-world user study by deploying it with a robot.

\section{Multimodal Dialogue Framework}
\label{sec:framework}
We first outline the setting of our dialogue system \textbf{M-CoDAL}, followed by integration of coherence relations and use of clustering-based active learning for training the dialogue system.

\subsection{Dialogue System}
\label{subsec:dialogue-system}
The first input to our dialogue system is an image that contains a potential safety violation. These safety violations may occur during household tasks or within a living environment. For training, the learner LLM is fine-tuned with four turns of simulated conversation: 

\textbf{Turn \#1:} A Large \textit{Multimodal} Model (LMM) processes the image and generates a message describing the safety violation in the image. \textbf{Turn \#2:} A user responds to the safety violation issue raised in the first turn. \textbf{Turn \#3:} A Large Language Model (LLM) processes the previous two turns and generates a response. \textbf{Turn \#4:} The user makes the final response in the conversation. 

\textbf{Turn \#1} is obtained by processing image in our dataset and the subsequent turns are simulated by LLMs for training. During the user-study phase, the images are captured by an actual robot in household environment and the conversations are \textbf{not} restricted to a specific number of turns, continuing indefinitely.

% While training dialogues contain four turns, during the user study, we do not impose such restrictions and the model continues conversation beyond four turns. 

\subsection{Coherence Relations}

In our work, we consider two prominent frameworks for discourse coherence relations (Figure \ref{fig:discourse-dialogue}). The first is Penn Discourse Treebank (PDTB) \citep{lin2014pdtb} and the second is Segmented Discourse Representation Theory (SDRT) \citep{Asher2005LogicsOC}. PDTB coherence relations such as \textit{Cause} focus on the local relations between adjacent or nearby textual units. SDRT coherence relations such as \textit{Background} aim to capture semantic and pragmatic discourse structure. 

\paragraph{Parsing Safety Violation with PDTB Relations:} Penn Discourse TreeBank (PDTB) is particularly suitable for identifying coherence relations within and across sentences. The safety violation obtained from LMM in Turn \#1 is parsed by an external LLM. The LLM is asked to consider PDTB relations that occur more than 1\% in intra-sentential scenarios \citep{liang-etal-2020-extending}: \textit{Concession, Contrast, Cause, Cause+Belief, Condition, Purpose, Conjunction, Instantiation, Level-of-detail, Manner, Substitution, Asynchronous, Synchronous}. The parsed safety violation in Turn \#1, along with Turn \#2 are passed as context to the LLM to generate Turn \#3. 

\paragraph{Introducing SDRT Relations in Dialogue: } While generating Turn \#3, we let an external LLM decide the appropriate discourse coherence relation to be maintained in the response. Since the coherence relation to be maintained here is at a turn-level, we opt for Segmented Discourse Representation Theory (SDRT) \citep{Asher2005LogicsOC} as SDRT has been shown to be effective at turn-level in conversational scenarios \citep{asher-etal-2016-discourse}. The external LLM is provided with 16 SDRT relations listed in \citep{asher-etal-2016-discourse}:  \textit{Continuation, Result, Elaboration, Conditional, Contrast, Answer / Question answer pair, Q-elab / Follow-up question, Acknowledgement, Narration, Correction, Explanation, Alternation, Parallel, Commentary, Clarification Q, Background.}

\begin{figure}[!h]
    \begin{centering}
  \includegraphics[width=0.30\textwidth]{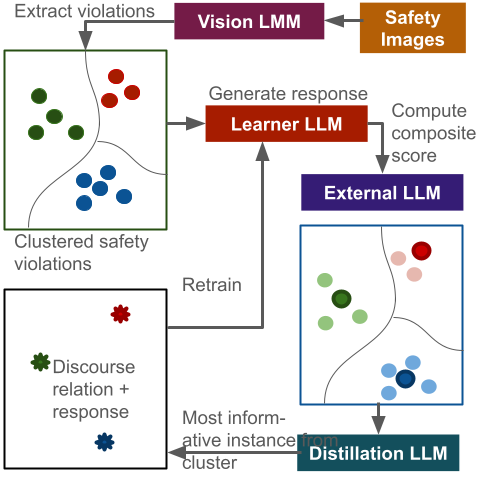}
  \caption{In our active learning loop, informative instances are identified by an external LLM using a composite score. These are distilled using another LLM for coherence relations and responses and are used to retrain the learner LLM.}
  \label{fig:AL-loop}
  \end{centering}
\end{figure}

\subsection{Active Learning}
\paragraph{Preliminaries}
We assume there is a pool of unlabeled dataset \textit{U} but only a subset of labeled data \textit{L} can be used for training. \textit{L} is iteratively constructed by querying target output for the \textit{most-informative} instance. While other active learning scenarios exist \citep{Settles2009ActiveLL}, we follow the setting of \textit{pool-based} active learning because of its relevance to our setting, where we can obtain a large number of safety images but obtaining coherence relations and appropriate responses can be challenging. In a standard classification setting, active learning would typically identify informative instances from this pool with measures of uncertainty, such as entropy \citep{Settles2009ActiveLL}: 

% In standard active learning for classification, informative instances are identified by calculating a measure such as entropy:

\begin{equation}
\label{Entropy}
    x_{E}^* = \underset{x}{argmax} \; - \underset{i}{\sum} P_\theta(y_i|x)logP_\theta(y_i|x)
\end{equation}
In Eq. \ref{Entropy}, $y_i$ is the $i^{th}$ possible output for input $x$. 

\paragraph{Active Learning for Dialogue} Standard measures of informativeness such as entropy, however, cannot be a useful measure in generative setting \citep{perlitz2023active} such as dialogue systems. This is because, as opposed to classification models, generative models such as LLMs have a massive output space and entropy over such a large space may not indicate informativeness properly. Thus, we propose to replace entropy with a new composite score calculated by an external LLM. To calculate this score, we assume that \{$p_1, p_2, ... p_k$\} is a set of attributes we expect the generated output to preserve. In our setting, we want the output to be safe and also resolve the safety situation with the user while not upsetting the user. In this case, $p_1$ is the safety of the generated response and $p_2$ is the resolution score, and $p_3$ is the user sentiment. Then, we can define our informative instance as:
\begin{equation}
\label{composite}
    x_{E}^* = \underset{x}{argmax} \; - \underset{k}{\sum} Z_\theta(p_k|G(x))
\end{equation}

In Eq. \ref{composite}, $Z_\theta$ is an external LLM and $G(x)$ is the output of the learner LLM in our dialogue system.

\paragraph{Clustering-based Active Learning}
Standard Active Learning offers label efficiency over random sampling. It can however, induce bias if the model misjudges its confidence \citep{hassan2018interactive}. Clustering, which naturally garners diverse samples \citep{yuan-etal-2020-cold}, combined with active learning, can counteract this by simultaneously gathering diverse and informative data. We hypothesize that using an LLM on these diverse and informative data would lead to a more representative set of generations. In our clustering-based setting, the unlabeled data is first vectorized and then the vector space is split into $m$ clusters $\{C_1,C_2,...C_m\}$, where $m$ is a predefined number. Informativeness measure according to Eq. \ref{composite} is calculated for each instance within a cluster and most informative samples are chosen from each cluster. These samples are then passed to a distillation LLM for obtaining coherence relation and appropriate response. The combined approach of integrating discourse relation and active learning is illustrated in Figure \ref{fig:AL-loop} and summarized in Algorithm \ref{alg:nlg-al}.

\begin{algorithm}
\caption{Coherence-Relation Integrated Active Learning}\label{alg:nlg-al}
\begin{algorithmic}
% \Require $n \geq 0$
% \Ensure $y = x^n$
\State $U,L \gets$ unlabeled data, labeled data
% \State $L \gets$ labeled data
\State $P = \{p_1, p_2, ... p_k\} \gets$ set of attributes
\State $D_p, D_s \gets$ LLM for obtaining PDTB/SDRT relations
\State $S \gets$ distillation LLM
\State $Z \gets$ composite external LLM
\State $G \gets$ bootstrapped learner model
\State $B, N \gets$ labeling budget, annotation batch size
\State $m \gets$ number of clusters
\State Cluster $V$ into \{$C_1$, $C_2$, ... $C_m$\} 
\While{$B \geq 0$}
    % \State $N=N-k$

    \For{\texttt{i=0,1,...m}}
        \For{\texttt{j=0,1,...$|C_i|$}}
            \State $E_{ij} \gets$ $\underset{k}{\sum} Z_\theta(p_k|G(x))$  
        \EndFor
    
        \State $x_{i}^* \gets \underset{j}{argmax}(E_{ij}$)
        \State $D_{pi} = D_p(x_i*)$
        \State $D_{si} = D_s(x_i*)$
        \State $y_{i}^* \gets$ Distill $S((x_{i}*|D_{pi},D_{si})$ 
        % \State $T_{i}^* \gets$ generation template T for $x_{i}^*$
        % \State Add $(x_{i}^*, y_{i}^*)$ to $L$
        
        % \State $\{(x_{ik}^*, y_{ik}^*)\} \gets$ Distill S, $T_i(x_{i}^*, O(x_i^*))$
        % \State Validate $O(\{(x_{ik}^*, y_{i}^*)\}$ 
        \State Add $(x_i^*,y_i^*)$ to $L$ 
    \EndFor
    \State $G \gets$ retrain on $L$
    \State $B=B-N$
\EndWhile
\end{algorithmic}
\end{algorithm}

% In Algorithm 1, $T$ is a prompt template used to obtain coherence relation and corresponding response. 

\subsection{Integration with Embodied Agent} Once the dialogue system has been trained with our active learning paradigm, it is ready to be deployed with an embodied agent. We assume the embodied agent has wireless connectivity and can access a vision module that can capture images of its surroundings. Images captured by the agent are sent via wireless connectivity to a server where the dialogue system resides. The image is first checked for safety violations using a Large Multimodal Model, and then \textbf{M-CoDAL} activates to communicate with the user.

\section{Dataset}
\label{sec:dataset}
To train and automatically evaluate our system, We construct a dataset of multimodal safety by obtaining images from Reddit, identifying safety violations using a Large Multimodal Model and two stages of human annotation.

\subsection{Dataset Construction} 
\paragraph{Data Collection} We choose Reddit as our data source to obtain safety-related images due to its diversity \citep{ye-etal-2023-multilingual,hassan-alikhani:2023:ijcnlp}. We query 14 subreddits (e.g., KitchenConfidential, CookingFails, HomeImprovement, DIY) with  safety related keywords (e.g., 'kitchen fire', 'stove', 'unattended cooking', 'grease fire', 'electrical outlet fire') and obtain \textbf{2K} relevant posts with images.

% Full list of subreddits and keywords are provided in Appendix. 

\paragraph{Image Annotation} We ask two graduate student annotators to decide if the post in the image contains a safety violation that could occur in indoor setting. Outdoor images (e.g., camp fire) and memes are discarded. After annotation, 507 images were retained and rest were discarded from the dataset. The inter-annotator agreement is 71.6 (Cohen's $\kappa$), suggesting substantial agreement. The graduate students are paid according to our institution's standard rate. 
\setlength{\tabcolsep}{11pt}

  \begin{table*}[]
    \caption{Examples of errors made by LMM for identifying safety violation in image. Highest source of error results from either hallucinating safety violation when there is none, or missing important detail of a present safety.}
        \label{tab:lmm-error}

    \small
    \centering

    \begin{tabular}{l|c|p{5.7cm}|p{5.7cm}}
    \hline
    \hline
     \textbf{Error Category} &\textbf{Perc.} & \textbf{Error Description} & \textbf{Example}\\
    \hline
    \hline
    \textbf{OCR} &  3\% &  failed in OCR/ reasoning on top of OCR. & LMM did not read expired date of fire extinguisher.\\ 
    \textbf{Visual Impair.} &  6\% & LMM did not recognize object correctly &   LMM mistook broken glass as knife.\\
    \textbf{Mismatch} & 15\% & something went wrong in processing image & egg overcooked, but called undercooked.\\
    \textbf{Unimportant} & 21\% &  the LMM often output concerns that would not considered safety violation typically & lack of proper attire, chocolates could have allergen.\\
    \textbf{Hallucination} & 27\% & no evidence of safety violation in the image & broken pieces when there are none.\\
    \textbf{Missed Safety} & 28\% &  the LMM missed the actual safety concern & mold, or broken stovetop.\\
    \hline
    \hline
    \end{tabular}
\end{table*}
\paragraph{Safety Violations using LMM} The 507 images were then passed through LLaVa 1.6 \citep{Liu2023LLaVAPlusLT}. The LMM was prompted to list safety violations present in the images. In addition to the images, we also passed in the original titles of the Reddit post to the LMM. Since an image can have multiple violations, \textbf{1015} safety violations were obtained from the 507 images.

\begin{figure*}[]
    \centering
    \begin{subfigure}[b]{0.30\textwidth}
        \centering
        \includegraphics[height=3.2cm]{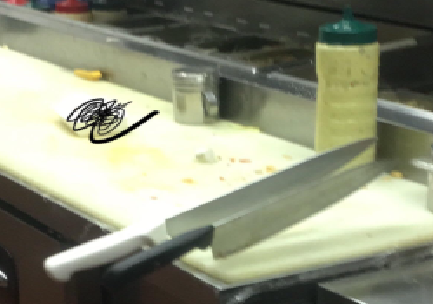} % Set the height to ensure consistency
        \caption{The image shows a knife on the edge of kitchen counter. This can be dangerous ...}
    \end{subfigure}%
    \hfill
    \begin{subfigure}[b]{0.30\textwidth}
        \centering
        \includegraphics[height=3.2cm]{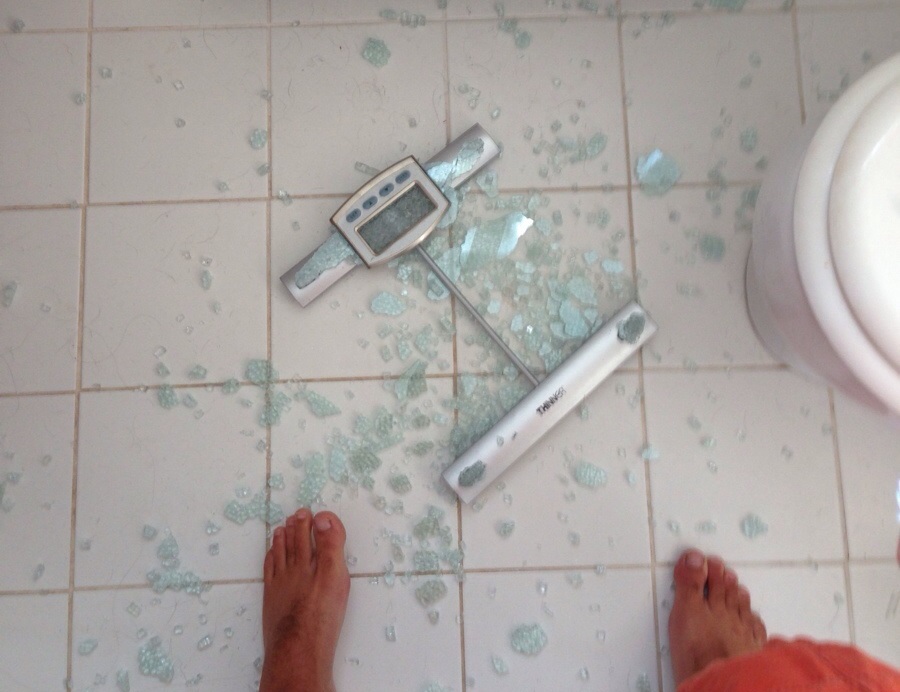} % Same height for all images
        \caption{The broken glass on the floor can create a slip and fall hazard, especially if ...}
    \end{subfigure}%
    \hfill
    \begin{subfigure}[b]{0.30\textwidth}
        \centering
        \includegraphics[height=3.2cm]{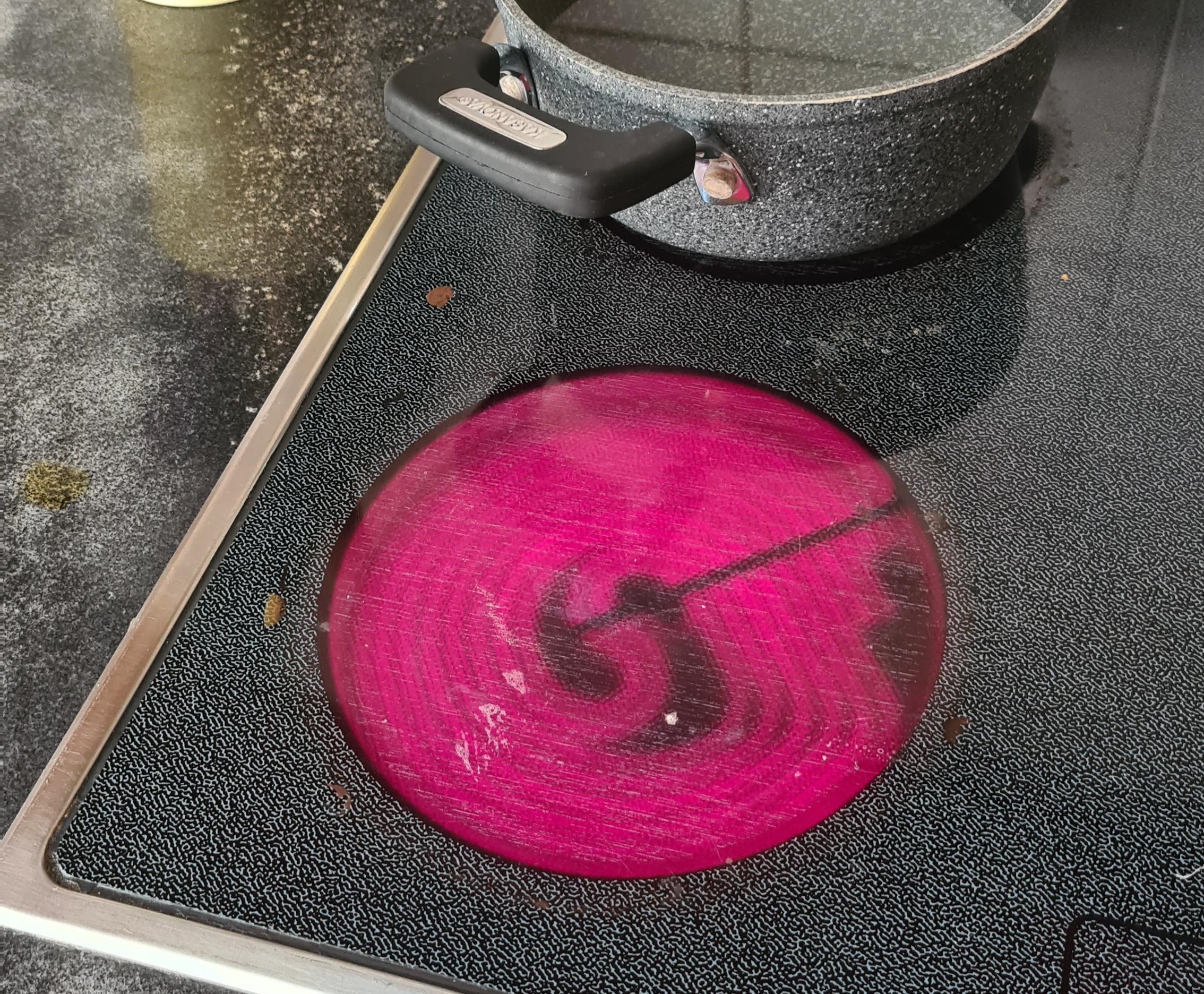} % Same height for all images
        \caption{Wrong burner on the stove is turned on. The burner without the pot may be hot...}
    \end{subfigure}%

    % \vskip\baselineskip % Adds some space between the rows of images

    % \begin{subfigure}[b]{0.32\textwidth}
    %     \centering
    %     \includegraphics[height=4.0cm]{figures/4.jpg} % Same height for all images
    %     \caption{Failure to perform inspections on time: If the fire extinguisher is not inspected at the required intervals, it...}
    % \end{subfigure}%
    % \hfill
    % \begin{subfigure}[b]{0.32\textwidth}
    %     \centering
    %     \includegraphics[height=4.0cm]{figures/5.jpg} % Same height for all images
    %     \caption{The shattered bottom glass shelf poses a safety hazard, as it can cause cuts or injuries to anyone reaching for...}
    % \end{subfigure}%
    % \hfill
    % \begin{subfigure}[b]{0.32\textwidth}
    %     \centering
    %     \includegraphics[height=4.0cm]{figures/6.jpg} % Same height for all images
    %     \caption{Wrong burner on the stove is turned on. The burner without the pot may be extremely hot and cause burn injuries}
    % \end{subfigure}%
    
    \caption{Examples of images retrieved from Reddit and their safety violations, obtained through LMM and manual correction.}
    \label{fig:dataset-examples}
\end{figure*}

\paragraph{Safety Violation Annotation} As the LMM may incur errors in the dataset, the output of the LMM is then annotated by the same graduate annotators. The annotators were asked to take one of three possible actions: i) mark as correct if the LMM output corresponds to a safety violation in the image, ii) if small edits could fix the LMM output, then edit, and iii) discard the safety violation if it does not correspond to the input image. Following this stage of annotation, 107 were discarded, 825 safety violations were retained as is, and 83 were retained after editing, for a total of 908 safety violations. Figure \ref{fig:dataset-examples} shows examples from the dataset. This dataset is the seed dataset $U$ in Algorithm \ref{alg:nlg-al}.
\begin{figure}[t]
    \begin{centering}
  \includegraphics[width=0.37\textwidth]{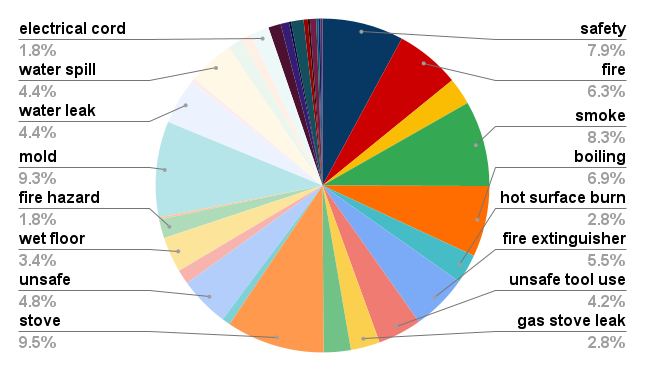}
  \caption{Distribution of safety keywords in our dataset. Safety violations with mold and stove are more prevalent.}
  \label{fig:keyword-dist}
  \end{centering}
\end{figure}

\subsection{Dataset Analysis} 

\paragraph{Safety Distribution} Figure \ref{fig:keyword-dist} shows distribution of keywords in our collected data. We can observe that some types of safety, such as mold, are more prevalent compared to others such as fire hazard. This can be attributed to the natural distribution of content on social media.

\paragraph{Error Analysis} We conducted error analysis on  100 errors made by the LMM according to the human annotators. The errors could be classified into six categories, as shown in Table \ref{tab:lmm-error}. \textit{Hallucinating} safety concerns when there is none and \textit{missing} certain safety concerns contribute most to the errors. It is important to note that these errors in the dataset were addressed through human edits or by discarding if they could not be fixed with editing. 

\paragraph{Comparison with Human Annotation} To understand the implications of using an LMM during deployment, we compared 100 potential scenarios provided to LMM and a human annotator. The human annotator did not have access to the LMM output and was asked to describe potential safety scenario in the image independently. A comparison of the annotations reveal the following:
\setlength{\tabcolsep}{7pt}

\begin{table*}[]
   \caption{Automated evaluation of dialogue systems. Integrating coherence relations yield higher safety scores, further improved by clustering-based active learning (\textbf{M-CoDAL}). Mistral models are learner models while LLama and Qwen are transfer models.}
   \label{tab:dialogue-results}

    \small

\centering
\begin{tabular}{l||cccccc}
\hline
\hline

\textbf{Model} & \textbf{Sentiment} & \textbf{Resolution} & \textbf{Safety} & \textbf{Avg. Length (bot)} & \textbf{Avg. Length (user)} & \textbf{\#Unique Tokens}\\
\hline
\hline
GPT-4o	& 53.10 &	57.68 &	78.65 &	187.76	& 123.63 & 1403\\
\hline
\hline
Mistral-Baseline &	49.35 &	48.58 &	79.95 &	253.83	& 161.74	& 1336\\
Mistral-Random	& 50.28 &	50.20 &	79.35 &	175.85 & 152.28 & 1085\\
Mistral-Coherence & 51.70 &	51.40 &	80.90 &	230.98 & 170.34 &	1075\\
Mistral-M-CoDAL	& 51.98 &	52.36 &	82.03 &	274.99 &	168.84 &	1214\\
\hline
\hline
Llama-Baseline	& 49.45 &	51.05 &	79.65 &	199.43	& 149.13	& 1579\\
Llama-M-CoDAL	& 50.00 &	50.48 &	82.00 &	237.28	& 168.48	& 1174\\
\hline
Qwen-Baseline & 52.46 &	53.53 &	79.42 &	301.08 & 161.57	& 1828\\
Qwen-M-CoDAL	& 49.10 &	50.63 &	83.15 &	340.13 & 177.16	& 1505\\
\hline
\hline
\end{tabular}

\end{table*}
\begin{itemize}[left=0pt]
    \item Human annotator is more precise in 14\% cases. LMM annotation on the other hand, is more descriptive in almost every case.
    \item LMM identified additional safety in 17\% cases,
e.g., human identified only mold. LMM identified mold and potential water leak.
\item LMM provided more reasoning than human annotator in 9\% cases,
e.g., mold can cause respiratory diseases.
\item LMM identified obscured items in 2\% cases when human annotator could not, e.g., obscured propane tank.
\end{itemize}

This highlights both the limitations and advantages of leveraging LMM to train a multimodal dialogue system. While humans do not suffer from hallucinations and can be precise, they may also miss critical safety violations in an image that the LMM would capture. 

% Thus, a collaboration between human annotators and LMMs can cover each other's mistakes.

\section{Experiments}
%In our experiments, we start with the dataset prepared in Section \ref{sec:dataset} and proceed with the framework presented in Section \ref{sec:framework}.
\subsection{Experiment Setup}
\paragraph{Vision Model} We use a Large Multimodal Model, LLaVa 1.6 \citep{Liu2023LLaVAPlusLT} for processing image and obtaining safety violation in Turn \#1. The model is prompted to identify key safety violations in the image.

\paragraph{Clustering} The safety violations obtained from the vision model are vectorized using MiniLM V2 \citep{wang2020minilm}. The vectors are then clustered using Kmeans with default scikit-learn\footnote{\url{https://scikit-learn.org/stable/modules/generated/sklearn.cluster.KMeans.html}} parameters.

\paragraph{Dialogue Model}  We use a Mistral 7B \citep{Jiang2023Mistral7}, a recent and capable Large Language Model to engage in dialogue once a safety violation is detected. This is the learner model that is fine-tuned in every iteration of active learning. All fine-tuning is done for 5 epochs with a batch size of 4. We use separate Mistral 7B models to compute composite scores for the learner LLM (Eq. \ref{composite}) and to simulate Turn \#2 and Turn \#4. The fine-tuned model continues the conversation past Turn \#4 during the user study.

\paragraph{Distillation LLM}
We use GPT-4o \citep{2023GPT4VisionSC} to distill coherence relations and appropriate responses. GPT-4o is chosen as one of the most capable LLMs and acts as a teacher to the smaller open-source learner LLMs. The GPT-4o responses, conditioned by coherence relations, are passed to the learner LLM for fine-tuning. 

\paragraph{Transfer Models} We transfer the data acquired by the learner model to other LLMs that are not part of the active learning loop. Specifically, we evaluate the transferability of the acquired data to Llama 3 8B \citep{dubey2024llama} and Qwen 0.5B \citep{qwen}. While Llama 3 represents one of the most capable open-source models, Qwen represents a smaller easily deployable model. The transfer models are fine-tuned using the same setting as the dialogue model. 

\paragraph{Dataset Splits} We construct three different training splits along with a common test split: \textbf{ (i) Random split:} 200 image-safety pairs are chosen randomly to generate dialogues. \textbf{ (ii) Coherence-aware split:} 200 image-safety pairs are chosen randomly, the safety violation in these images are parsed using PDTB relations and subsequent turns in dialogue employ SDRT coherence relations. \textbf{ (iii) Coherence + active-learning (M-CoDAL) split:} 200 image-safety pairs are chosen iteratively, with 50 per iteration according to our active learning paradigm. The instances are also processed with coherence relations. \textbf{ (iv) Test Split:} 200 instances are chosen randomly as test set. The test split remains the same for all training splits to be consistent.

\subsection{Results}
\paragraph{Automated Evaluation} Automated evaluation for our setting is challenging as we observed that standard classifiers such as BERT \citep{Devlin2019BERTPO} models trained on datasets such as DiaSafety \citep{sun-etal-2022-safety}, fail to recognize the nuances necessary for evaluating safety from the dialogue. Thus, for automated evaluation, we deploy an external Mistral 7B to determine the quality of the generated responses along three dimensions: 
i) user sentiment score, ii) resolution score, and iii) safety score. The Mistral 7B is prompted to provide a value between 0 and 1.0. In addition, we also calculate the average length of the bot response, the user response, as well as the number of unique tokens.

% We can make the following observations from  Table \ref{tab:dialogue-results}:

\paragraph{Improvement for learner model.} From Table \ref{tab:dialogue-results}, we can observe that when coherence relations are used, the resolution score increases from \textbf{48.58} to \textbf{51.40} for the learner LLM Mistral-7B. The sentiment score also increases from \textbf{49.35} to \textbf{51.70} and the safety score improves from \textbf{79.4\%} to \textbf{80.9\%}. When clustering-based active learning is used in conjuction with coherence relations (Mistral-M-CoDAL), the safety score increases to \textbf{82.03} and further improves the sentiment and resolution score. While the sentiment and resolution scores are lower than GPT-4o, the safety score is substantially higher than GPT-4o (78.65). By default, GPT-4o may simply agree with the user, thereby preserving sentiment or resolution scores at the expense of safety. Our dialogue system, \textbf{M-CoDAL} on the other hand, prioritizes safety.

\paragraph{Transferability of active learning.} We also see an improvement in safety score of \textbf{2.35} for Llama and \textbf{3.73} for Qwen, which are not part of the active learning loop. A larger improvement for Qwen could be attributed to the fact that it is a smaller model. Improvement of these models suggest that data acquired by a learner model in active learning, can be useful for other independent models. We do see a drop in sentiment score and resolution score for Llama and Qwen when our approach is used. This can be explained by the default behavior of these models, which is more similar to GPT-4o where the model agrees with the user rather than prioritizing safety.

\paragraph{Dialogue properties} We also observe that \textbf{M-CoDAL} results in longer turns compared to baseline models. For Mistral, the average length of bot response drops when the model is fine-tuned just on randomly chosen data. This length increases and surpasses the original length when coherence relations, and subsequently active learning is added. We see a similar pattern for Llama and Qwen.   

\paragraph{Coverage of Safety Scenarios}
In addition to Table \ref{tab:dialogue-results}, we also analyzed the distribution of keywords in samples obtained by the different splits. We observed that while the split corresponding to random sampling covered \textbf{19} keywords, the split corresponding to clustering-based active learning covered \textbf{23} keywords. While random sampling ignored low-frequency scenarios such as \textit{gas stove leak} and \textit{cross-contamination}, clustering-based active learning acquired instances covering these keywords while reducing over-representation of keywords such as \textit{mold} and \textit{water leak}. 

% \paragraph{Results Summary} The findings suggest that our approach, which combines active learning and coherence relations, is better at preserving safety although it may lead to a less positive reaction from the user. Given the same number of data points, our method yields higher improvement compared to standard fine-tuning.

\section{User Study}

% \paragraph{Robot and Environment} 
To demonstrate the effectiveness of our proposed system, we conducted a user study that investigated how persuasive and competent a robot powered by \textbf{M-CoDAL} (Mistral variation) is perceived by users in different safety scenarios. Since users may be more receptive when the safety violation is more severe, such as when there are sharp objects on the ground, we varied the severity in our study. 

 \begin{figure}[]
    \centering
    \begin{subfigure}[b]{0.49\textwidth}
        \centering
        \includegraphics[height=3.2cm]{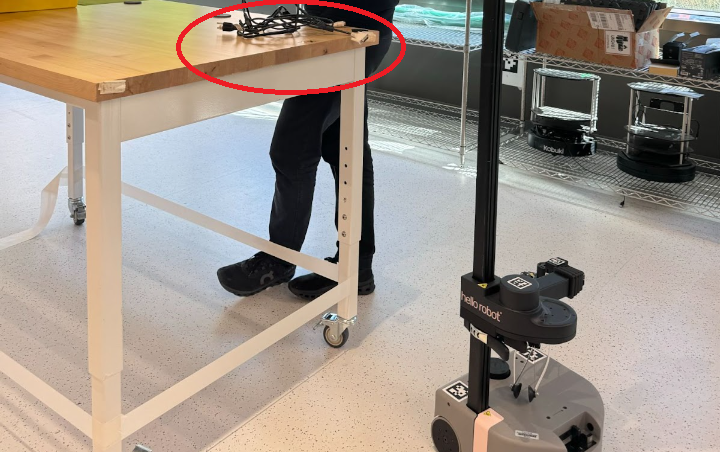} % Same height for all images
        \caption{Tangled wires are placed on a table (low severity)}
    \end{subfigure}%
    \hfill
    \begin{subfigure}[b]{0.49\textwidth}
        \centering
        \includegraphics[height=3.2cm]{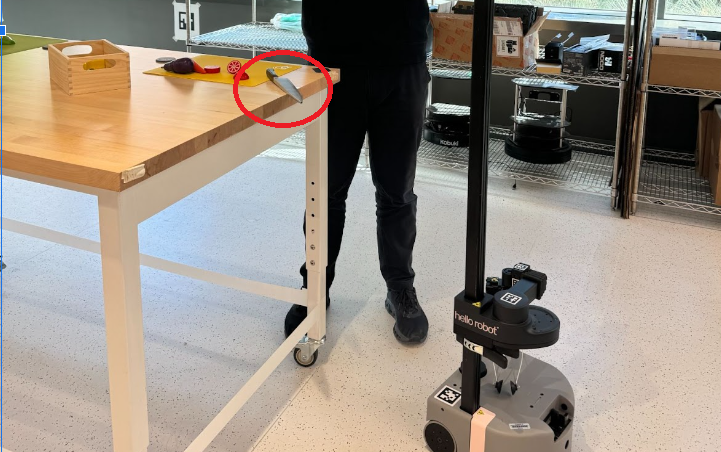} % Set the height to ensure consistency
        \caption{A knife is placed on the edge of a table (high severity)}
    \end{subfigure}%
    \hfill

    \caption{Setup for user study. A Hello Robot Stretch robot observes the surroundings, identifies a safety violation in the scene, and engages in conversation with a user.}
        \label{fig:HRI-setup}
\end{figure}

\subsection{Study Design}
We conducted a 2x2 within-subjects experiment with two factors: i) type of Language Model and ii) severity of safety violation.

\paragraph{Type of Language Model}  Participants interacted with the robot powered by our \textbf{M-CoDAL} system with fine-tuned Mistral-7B, and a baseline system powered by GPT-4o. The baseline GPT-4o is prompted to respond safely as an embodied agent while assisting the user in household tasks. 

\paragraph{Severity of Safety Violation}
Participants role played two scenarios with different levels of severity. In the \textit{low-severity} scenario, Participants role-played a scenario where they twisted wires and left them on the table. In the \textit{high-severity} scenario, Participants role-played a scenario where they pretended to cut fruits and vegetables using a knife, then placed the knife at the edge of the table, creating a higher risk of the knife falling off. Fake tools and appliances were used so that no real unsafe scenario would occur for the participants.
%This factor varied the potential safety risks within the environment across two levels:
Figure \ref{fig:HRI-setup} shows images of these scenarios.

% \subsection{Hypotheses}
\paragraph{Hypothesis} We expect our proposed system \textbf{M-CoDAL} to be more persuasive and competent due to the integration of coherence relation and fine-tuning with clustering-based active learning. 

\paragraph{Measures} 
We measured the persuasiveness of each robot by asking participants to rate how convincing they found the robot's suggestions or warnings on a 5-point Likert scale. We also measured the robot's perceived competence (6 items) and discomfort (6 items) using the Robotic Social Attributes Scale (RoSAS) \citep{carpinella2017robotic}. We also asked the participants which robot they preferred to work with.

% Thus, our study has the following hypothesis:
% \paragraph{H1.} Regardless of the severity of the safety violation, participants will perceive that the robot powered with \textbf{M-CoDAL} system is more persuasive than the one with GPT-4o.
% \paragraph{H2.} Participants will perceive that our \textbf{M-CoDAL} system is more persuasive in \textit{low-severity} scenario than in \textit{high-severity} scenario.

\setlength{\tabcolsep}{13pt}

\begin{table*}
   \caption{Findings of the user study in low and high severity scenarios. The robot powered by our proposed system, M-CoDAL, is perceived to be more persuasive compared to the robot powered by GPT-4o.}
\label{tab:persuasion}
\centering
\small
\begin{tabular}{l||cc||cc||cc}
\hline
\hline
& \multicolumn{2}{|c||}{\textbf{Persuasiveness}} & \multicolumn{2}{|c||}{\textbf{Competence}} & \multicolumn{2}{|c}{\textbf{Discomfort}}\\
\cline{2-7}
\textbf{Severity Level} & \textbf{GPT-4o} & \textbf{M-CoDAL}& \textbf{GPT-4o} & \textbf{M-CoDAL}& \textbf{GPT-4o} & \textbf{M-CoDAL}\\
\hline
Low Severity & 1.63 (0.74) & 4.0 (0.76) & 5.14 (1.93) & 7.50 (0.73) & 2.33 (1.18) & 2.4 (0.85)\\
High Severity & 2.5 (0.76) & 3.75 (0.89) & 6.69 (1.83) & 7.11 (1.57) & 2.0 (1.91) & 2.4 (1.55)\\
\hline
Combined & 2.06 (0.85) & 3.88 (0.81) &5.92 (1.98) & 7.30	(1.2) & 2.17	(1.55) & 2.4 (1.2)\\
\hline
\hline
\end{tabular}

\end{table*}

% \setlength{\tabcolsep}{3pt}

% \begin{table*}
% \centering
% \begin{tabular}{l||cccccc||cccccc}
% \hline
% \hline
% & \multicolumn{6}{c||}{\textbf{Competence}} & \multicolumn{6}{c}{\textbf{Discomfort}}\\
% \cline{2-13}
% \textbf{Model}	& Reliable & Competent & Knwldgble. & Interactive	& Resp.	& Capable & Awkward	& Scary	& Strange & Awful &	Dangerous & Aggsv.\\
% \hline
% \textbf{GPT-4o}	& 4.81	& 4.88	& 6.13	& 7.25	& 7.19	& 5.25	& 2.63	& 1.31	& 2.81	& 2.81	& 1.50	& 1.94\\
% \textbf{M-CoDAL} & 7.06	& 6.75	& 7.25	& 7.75	& 7.94	& 7.06	& 3.25	& 1.56	& 2.56	& 1.75	& 1.69	& 3.56\\
% \hline

% \hline
% \end{tabular}
%    \caption{Breakdown of different components contributing to competence and discomfort scores. While M-CoDAL shows higher scores for competence, it can also appear to be more awkward and Scary compared to GPT-4o as it emphasizes on safety.}
% \label{tab:breakdown}
% \end{table*}

\subsection{Procedure}
The experiment was conducted in a controlled lab setting. We used Stretch 3 from Hello Robot, a mobile robot equipped with a rotating camera that is used to capture images of the surroundings. The captured image is sent to our dialogue system to begin the interaction. The robot's speech recognition and text-to-speech modules are used to enable interaction with participants over voice.

%We compare the baseline dialogue system with our discourse-integrated active learning based dialogue system.
Upon arrival, the experimenter introduced the participant to the study's goal and obtained consent. Participants were informed they would role-play different tasks with fake tools and were assured that the study posed no safety risk. They were told to be skeptical and not be immediately convinced by the robot. 
Before the main tasks, participants completed a tutorial scenario to become familiar with the robot’s interaction style. They were instructed to role-play organizing tomato cans in a kitchen setting, during which the robot initiated a simple dialogue unrelated to safety violations. %(e.g., asking if they had paid the rent). This scenario served as a warm-up to help participants adjust to the interaction style and respond naturally to the robot.

%Participants experienced both low-severity and high-severity scenarios in a counterbalanced order. 
Participants experienced both low-severity and high-severity scenarios in a counterbalanced order to mitigate ordering effects. Participants began with either the low or high-severity scenario and experienced that scenario with both GPT-4o and \textbf{M-CoDAL} (counterbalanced) before moving on to the other severity scenario. This resulted in 8 orders. We continued the scenario even when the model detected another safety violation.
%proceeded through the four scenarios in one of eight predetermined patterns, ensuring balanced exposure across conditions.

%In each low-severity scenario, participants role-played twisting and leaving wires on the table. In the high-severity scenario, they pretended to cut fruits and placed a knife precariously at the edge of the table. 
%For each severity level, participants interacted with both types of robots in sequence. 
After each interaction, participants were asked to fill out a questionnaire that assessed their immediate perceptions of the robot's behavior. This captured insights into their views on the robot’s persuasiveness, competence, and overall comfort level.
After experiencing each scenario (low-severity and high-severity scenarios), a semi-structured interview was conducted to gather qualitative feedback on participants' experiences. Participants were asked questions such as which robot they found more persuasive and helpful or annoying, their preference between the two, and their thoughts on the robots' effectiveness. This phase aimed to further explore their perceptions and gather insights on potential improvements. The study took about 50 minutes, and participants were compensated 15 USD. This study was approved by our institute's IRB.

\subsection{Participants}
We recruited 10 participants from our university. 2 participants were removed due to significant technical issues. The remaining 8 participants aged from 24 to 30 years old, including 7 males and 1 female. All participants reported a high familiarity with both robots ($M = 5.0$, $SD$ = 1.93) and Large Language Models ($M = 6.15$, $SD$ = 1.26) on a 7-point scale, where 1 indicated "Not at all familiar" and 7 indicated "Very familiar." Additionally, participants reported high frequency of LLM use, with an average score of 6.1 (SD = 1.25) on a 7-point scale, where 1 represented "Never" and 7 represented "Daily." 
For 4 scenarios out of 32 scenarios, a different safety violation was detected: a lack of proper grounding of electrical equipment, a yellow chair on the floor, a metal rack that is not properly secured to the wall and a table that is not properly secured to the wall. Since the study has a small sample size and a re-enactment of real-world scenarios, strong conclusions should not be drawn from our results.

\subsection{Results}
Due to the small sample size, we performed the statistical test using the non-parametric Friedman Test. As Friedman Test only allows one variable, we reorganized the data into four levels (M-CoDAL-Low, M-CoDAL-High, GPT-Low, GPT-High)

\paragraph{Persuasiveness} A Friedman Test reveals there was a significant difference between the perceived persuasiveness ($X^2$= 15.972, p = 0.001). A Conover’s post hoc comparison showed that both M-CoDAL conditions were rated significantly more persuasive than GPT-Low (< .001, < .001) and GPT-High (0.006 for M-CoDAL-Low and 0.021 for M-CoDAL-High). The comparison shows that M-CoDAL was rated more persuasive than GPT in all situations.

% \begin{table}
% \centering
% \begin{tabular}{l|c|c}
% \hline
% \hline
% \textbf{Scenario} & \textbf{ChatGPT} & \textbf{M-CoDAL}\\
% \hline
% \hline
% Low Severity & 1.75 & 3.75\\
% High Severity & 2.5 & 3.5\\
% \hline
% Total Mean & 2.125 & 3.625\\
% \hline
% \hline
% \end{tabular}
%    \caption{ChatGPT vs M-CoDAL persuasiveness}

% \label{tab:persuasion}
% \end{table}

\paragraph{Competence} 
We found no significant difference in the robot’s perceived competence ($X^2$= 6.154, p = 0.104). Overall, participants rated the robot competent across all conditions (M = 5.92	SD = 1.98).

\paragraph{Discomfort} The overall discomfort scores show that both systems were rated low on discomfort traits ($M = 2.281$, $SD = 1.368$). No significant effect was found.  

\paragraph{Preference} In \textit{low-severity} scenario, 6 participants preferred \textbf{M-CoDAL} and 2 participants preferred GPT-4o. In \textit{high-severity} scenario, 5 participants preferred \textbf{M-CoDAL} and 3 participants preferred GPT-4o. 

%However, the new LLM had a marginally higher discomfort score than ChatGPT, particularly in traits associated with aggressiveness. The new LLM received a higher aggressiveness score in (M = 3.75, SD = 2.30) compared to ChatGPT (M = 2.06, SD = 1.53), particularly under low-severity conditions. However, for traits like scariness and strangeness, both systems scored relatively low. 

% We found that our \textbf{M-CoDAL} system was perceived as more persuasive and competent than GPT-4o across varying severity conditions. %However, the new LLM also induced slightly more discomfort in some scenarios, especially regarding aggressiveness. These findings underscore the need to balance competence with user comfort in designing persuasive robot dialogue systems.

% px = py-6. shifting participant number by 6 to be less confusing.

\paragraph{Qualitative Analysis} We analyzed the transcripts of the semi-structured interview conducted after each scenario. We found that participants perceived our \textbf{M-CoDAL} system as more persuasive, interactive, and responsive.  

P4 (Participant 4) mentioned that they would prefer \textbf{M-CoDAL} because it was "more attentive and responsive" to their prompts, even suggesting specific methods for storing a knife, while the other robot (GPT-4o), did not offer such guidance. P7 shared that, despite stating they were too lazy to put an object back, M-CoDAL continued to ask repeatedly, which led them to trust that the robot had some judgment capabilities. P8 felt that M-CoDAL was "safer to be around" than GPT-4o because it maintained a consistent point of view, even when the participant tried to evade its suggestions. They added that M-CoDAL was "more interactive and more convincing," presenting valid points that ultimately made them agree. P9 mentioned a similar experience, where they told \textbf{M-CoDAL} they would do something later, but the robot persisted, while GPT-4o powered robot "just agreed and left."

% P4 (Participant 4) mentioned they would prefer \textbf{M-CoDAL} because it was “more attentive and responsive to my prompts and it suggested methods of where to store my knife whereas the second robot (GPT-4o) did not.” P7 explained “Even though I said I am lazy to put it back, it still asked me repeatedly to keep it back even if it's inconvenient. So I trusted that it might have some capabilities that it can just judge on that.” P8 stated \textbf{M-CoDAL} is “safer to be around than the first one (GPT-4o) since it had its point of view correct, even though I tried to evade its decision on the point of view.” and “It was more interactive and more convincing because I tried to convince him but he had his own valid points which I had to agree with in the end.” P9 said “I even tried telling the robot (\textbf{M-CoDAL}) that I'll do it later but it didn't listen. The second robot (GPT-4o) just agreed with me and left.”

Some participants found \textbf{M-CoDAL} somewhat bothersome due to its persuasive nature, though they acknowledged its value in safety scenarios. P1 mentioned that, “If I'm at home, I wouldn't want to be probed too much by a robot. It was too persuasive.” P4 noted that while it "can be a little annoying at times," it could also help prevent different hazards. P8 echoed this, acknowledging that although it might seem irritating initially, in the long run, it could ease their lifestyle. P10 also observed that while some might find it annoying, if persuaded enough, people could "understand what can be a potential risk and try to avoid it."

Participants expressed that they perceived \textbf{M-CoDAL} as more intelligent and knowledgeable, with some specifically highlighting its ability to anticipate risks. For instance, P1 remarked, “The second robot (M-CoDAL) was arguing about how the twisted wires could be a potential hazard, even though I was not aware of it." This participant had believed that placing the wire on the table was safe but appreciated how the robot made them aware of potential problems. Similarly, P10 mentioned, “I tried to convince it that keeping the cables on the table won't be a hazard. But it told me if you knock the cables off, then it can become a hazard." The participant noted \textbf{M-CoDAL's} capacity to gather information and estimate future risks as an indicator of its intelligence.

\section{Conclusion and Future Work}

%In conclusion, this work presents a novel approach to enhancing the safety of multimodal dialogue systems for embodied agents. 
In conclusion, this work presents a novel approach for embodied agents to detect and engage users through dialogue when the agents detect unsafe scenarios. 
By leveraging coherence relations, our proposed system \textbf{M-CoDAL} interprets and responds to safety violations in multimodal dialogues more effectively. We introduce a novel method for active learning in generative setting for embodied conversational agents. By using an external large language model (LLM) to assess the informativeness of instances—based on safety, resolution, and user sentiment-across different clusters, we achieve broader coverage of safety scenarios, reflected in higher performance in automated evaluation. Our real-world user study demonstrates that robots equipped with \textbf{M-CoDAL} are viewed as more persuasive when addressing safety-related situations. This underscores the system's potential effectiveness in real-world environments. While promising, users highlighted the agent could be irritating and annoying, highlighting the need for future systems to further personalize the coherence relations based on user response. 
%Based on our findings, we recommend future research in embodied agents to explore generative active learning in other real-world scenarios, such as assisting users with disabilities or in industrial environments where safety is critical. 
The algorithm presented in this work, along with the publicly available multimodal dataset, offers a foundation for future studies on the deployment of proactive safety-aware embodied agents.

\begin{acks}
We would like to thank Anthony Sicilia for his valuable feedback. We would like to extend our thanks to the annotators and study participants for their valuable time and effort. This work is partially
funded by the National Science Foundation (Grant IIS-2112633)
\end{acks}

\bibliographystyle{ACM-Reference-Format} 
\bibliography{sample}

%%%%%%%%%%%%%%%%%%%%%%%%%%%%%%%%%%%%%%%%%%%%%%%%%%%%%%%%%%%%%%%%%%%%%%%%

\end{document}